\documentclass[a4paper,conference]{IEEEtran}
\usepackage{amsmath,amssymb}
\usepackage{url}

\usepackage{graphicx}
\usepackage{xspace}
\usepackage{csquotes}
\usepackage{cleveref}
\usepackage{booktabs}
\usepackage[caption=false]{subfig}
\newcommand{\etal}{\emph{et al.}\xspace}
\newcommand{\eg}{\emph{e.g.,}\xspace}
\newcommand{\ie}{\emph{i.e.,}\xspace}

\begin{document}
%
\title{Making Every Label Count: Handling Semantic Imprecision by Integrating Domain Knowledge}

\author{\IEEEauthorblockN{Clemens-Alexander Brust and Björn Barz and Joachim Denzler}
	\IEEEauthorblockA{Computer Vision Group\\Friedrich Schiller University Jena\\
		Jena, Germany\\
		Email: \{clemens-alexander.brust,bjoern.barz,joachim.denzler\}@uni-jena.de}}

\maketitle

\begin{abstract}
	Noisy data, crawled from the web or supplied by volunteers such as Mechanical Turkers or citizen scientists, is considered an alternative to professionally labeled data.
	There has been research focused on mitigating the effects of label noise.
	It is typically modeled as inaccuracy, where the correct label is replaced by an incorrect label from the same set.
	We consider an additional dimension of label noise: imprecision.
	For example, a \emph{non-breeding snow bunting} is labeled as a \emph{bird}.
	This label is correct, but not as precise as the task requires.
	
	Standard softmax classifiers cannot learn from such a weak label because they consider all classes mutually exclusive, which \emph{non-breeding snow bunting} and \emph{bird} are not.
	We propose CHILLAX (Class Hierarchies for Imprecise Label Learning and Annotation eXtrapolation), a method based on hierarchical classification, to fully utilize labels of any precision.
	
	Experiments on noisy variants of NABirds and ILSVRC2012 show that our method outperforms strong baselines by as much as 16.4 percentage points, and the current state of the art by up to 3.9 percentage points.
\end{abstract}


%
\IEEEpeerreviewmaketitle

\section{Introduction} 
When acquiring labeled training data for a budget-constrained image classification application, there is a trade-off between quantity and quality of the data.
Combined annotations from multiple domain experts can be considered correct, but are very expensive.
Data from automatic web crawling or volunteers is more cost-effective but noisy.

\begin{figure}
	\centering
	\includegraphics{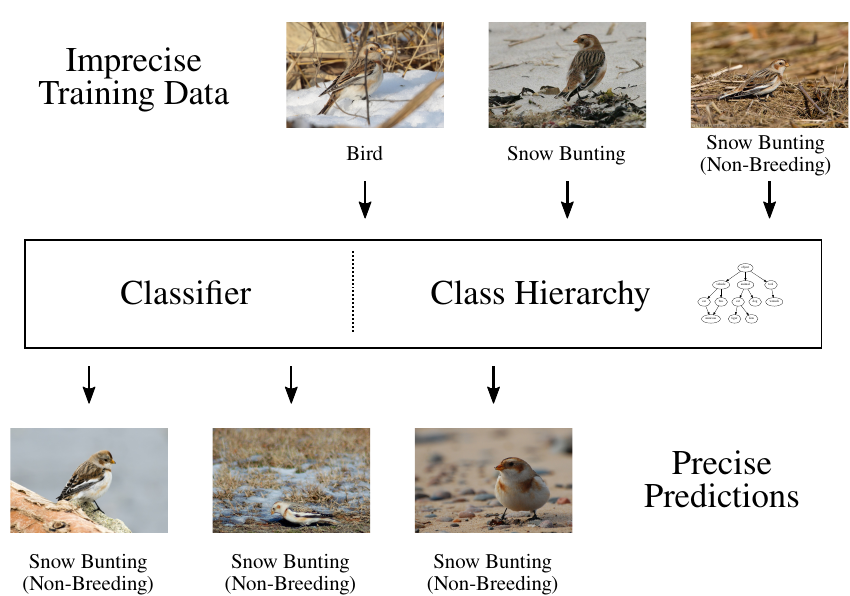}
	\caption{Learning from imprecise data. Our method CHILLAX is capable of fully utilizing training data where labels are imprecise, \ie not labeled as leaf nodes of a given class hierarchy. Predictions are always precise.}
	\label{fig:teasertop}
\end{figure}

Learning from noisy training data has thus been an important research topic.
However, most work approaches label noise from only one angle.
Labels are either correct or not, in which case they are confused with another label. The additional component of label \emph{precision} is usually not taken into account.
For example, a person might annotate an image of a North American barn swallow as simply a perching bird. This label is not inaccurate since it is correct, but it is imprecise.
Learning from such an imprecise label is a case of weakly supervised learning, which generalizes the trade-off between annotation quality and quantity.
However, unlike training labels, all predictions are expected to be precise.
In real-world applications, an output of \enquote{bird} is not helpful to the end user.
We visualize this scenario in \cref{fig:teasertop}.

In this work, we investigate the implications of imprecisely labeled data.
Because standard softmax classifiers are incompatible with imprecise labels, we propose our own method CHILLAX (Class Hierarchies for Imprecise Label Learning and Annotation eXtrapolation).
It is based on hierarchical classification.
This choice has two benefits:
\begin{itemize}
	\item The classifier can utilize all of the available data.
	      Conventional softmax classifiers can only learn from examples labeled at leaf nodes of the class hierarchy, because the rest violates their assumption of mutually exclusive classes. Such violations occur in large web crawled datasets like ImageNet~\cite{deng_imagenet:_2009}. It contains the concepts \enquote{electronic device} and \enquote{video display} which each consist of some unique images but are not mutually exclusive.
	\item We integrate domain knowledge in the form of a class hierarchy directly into the classification process.
	      This additional information can improve accuracy and learning speed \cite{brust_integrating_2019}.
\end{itemize}

To evaluate the performance of our method, we build upon on the well-known benchmark datasets NABirds \cite{van_horn_building_2015} and ILSVRC2012 \cite{russakovsky_imagenet_2015}.
We reduce the label precision of this data synthetically to simulate imprecisely labeled data.
The reduction is performed using statistical models for label precision depending on the specific data source.
One example is web crawling for training data, or webly supervised learning, a research area of increasing relevance.
Volunteer annotators such as citizen scientists or hobbyists are considered as well.
We construct the statistical models to reflect the unique characteristics of these data sources.
Furthermore, we validate our noise models on image metadata obtained by web crawling.
This shows that our synthetic data matches the properties of real-world data.

Experiments on this synthetic data compare CHILLAX against strong baselines and the state of the art HEX~\cite{deng_large-scale_2014}.

\section{Related Work}
\label{sec:relatedwork}
While the topic of semantically imprecise labels is not widely considered, there is still relevant literature in closely related fields.
This section gives a brief overview of work concerning models of label noise and hierarchical classification methods.

Label noise is usually described in terms of confused labels. Patrini \etal model label noise as class-conditional \cite{patrini_making_2017}.
Each class has a certain probability of being changed to any other class. These probabilities are defined using a transition matrix.
It could describe confusions caused by similarity, especially in a fine-grained setting.
In \cite{li_learning_2017}, label noise is modeled more generally.
A completely unknown process changes the labels, \ie it may not even depend on the ground truth.
In \cite{dehghani2017learning}, label noise is considered in a weakly supervised learning setting.
These works do not consider semantics explicitly.
However, there is evidence that uniformly random confusions are not representative of real-world label noise~\cite{jiang2020noise}, highlighting the importance of our investigation and the relations between labels.
Also, \cite{qin2004tree} offers a semantic perspective on uncertainty as \enquote{fuzzy sets}. Here, the imprecision is not in the annotations, but the class labels themselves.

Semantics are looked at more often for improving learning methods.
For example, \cite{bertinetto_making_2019} discusses the benefits of a hierarchical perspective on classification.
However, most methods that leverage class hierarchies are used for metric learning \cite{hwang_learning_2011,verma_learning_2012,zhang_embedding_2016}, zero-shot learning  \cite{huo_zero-shot_2018,li_zero-shot_2015,rohrbach_transfer_2013,ye_zero-shot_2017}, as a regularizer \cite{fergus_semantic_2010,goo_taxonomy-regularized_2016,srivastava_discriminative_2013}, or to build an embedding space \cite{barz_hierarchy-based_2019,faghri_vse++:_2017,frome_devise:_2013,hwang_unified_2014}.
Deng \etal acknowledge the problem of semantic imprecision as a result of crowdsourced or crawled annotations in \cite{deng_large-scale_2014}.
Yet, their model of label noise is very simplistic as it only looks at the bottom two layers of a class hierarchy consisting of more than ten layers.
It therefore fails to capture the nuances of differently skilled human annotators (see \cref{ssec:dengmodel}).
The notion of label precision is balanced with accuracy on a prediction level in \cite{deng2012hedging}.

We build our method CHILLAX upon the work by Brust \& Denzler \cite{brust_integrating_2019}.
They propose a probabilistic hierarchical classifier to improve accuracy over conventional softmax classifiers.
A probabilistic model encodes common assumptions about class hierarchies.
To apply it to deep classifiers, they derive a label embedding and a task-specific loss function.
In the following, we will show how this method can be adapted to accommodate imprecisely labeled data.

\section{Learning from Semantically Imprecise Data}
\label{sec:sid}
This section details our definition of semantically imprecise data and the construction of our method for fully utilizing it, called CHILLAX.

Let $f: \mathcal{X} \to \mathcal{Y}$ be a classifier with:
\begin{itemize}
	\item $\mathcal{X}$ the input images, \eg $\mathbb{R}^{224\times 224}$.
	\item $\mathcal{Y}$ the label set, \eg the set of all bird species.
\end{itemize}

We then define the \emph{imprecise} label set $\mathcal{Y}^+$ as the set of all semantic ancestors of $\mathcal{Y}$.
In the example of bird species, the set $\mathcal{Y}^+$ corresponds to regna, phyle, familiae etc.
For better contrast, we call $\mathcal{Y}$ the \emph{precise} label set.

We require the classifier to predict only precise labels from $\mathcal{Y}$.
At the same time, it needs to be able to learn from training data with labels from both $\mathcal{Y}$ and $\mathcal{Y}^+$, which we refer to as semantically imprecise data.
This formulation is a weakly supervised learning scenario.
However, the training set may contain both weak (imprecise) and strong (precise) labels at the same time.

\subsection{Hierarchical Classification}

\begin{figure*}[t]
	\centering
	\subfloat [Original loss mask from \cite{brust_integrating_2019}. This annotation of \enquote{animal} is interpreted as neither \enquote{cat} nor \enquote{dog}. The respective nodes are trained to $p=0$.]{%
		\includegraphics[width=0.40\linewidth]{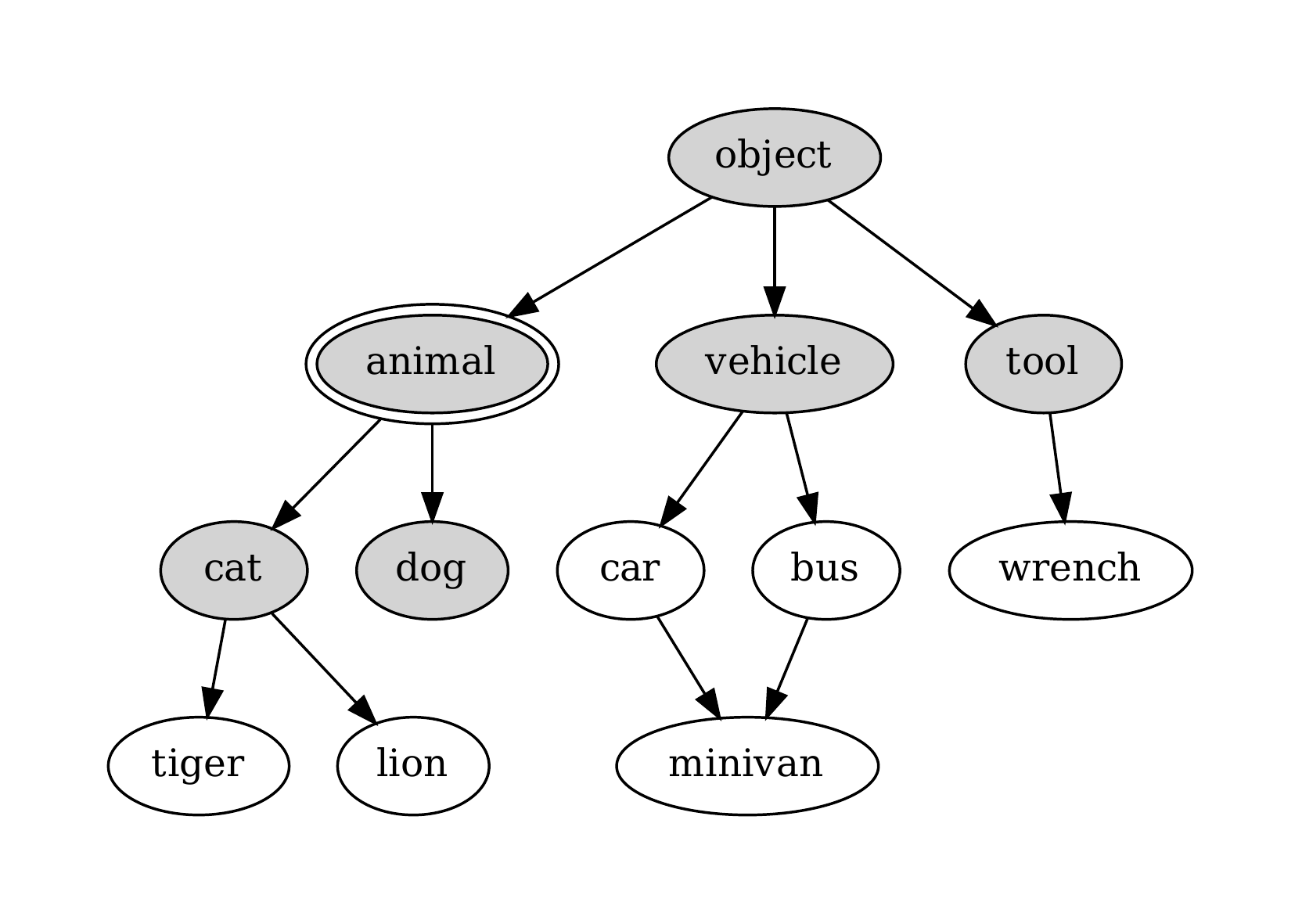}
		\label{fig:mdlmaskorig}
	}\hfil
	\subfloat [CHILLAX loss mask. With an imprecise \enquote{animal} annotation, it could be a \enquote{cat} or \enquote{dog}, but it is unknown which. The respective nodes are not trained to reflect this lack of knowledge.]{%
		\includegraphics[width=0.40\linewidth]{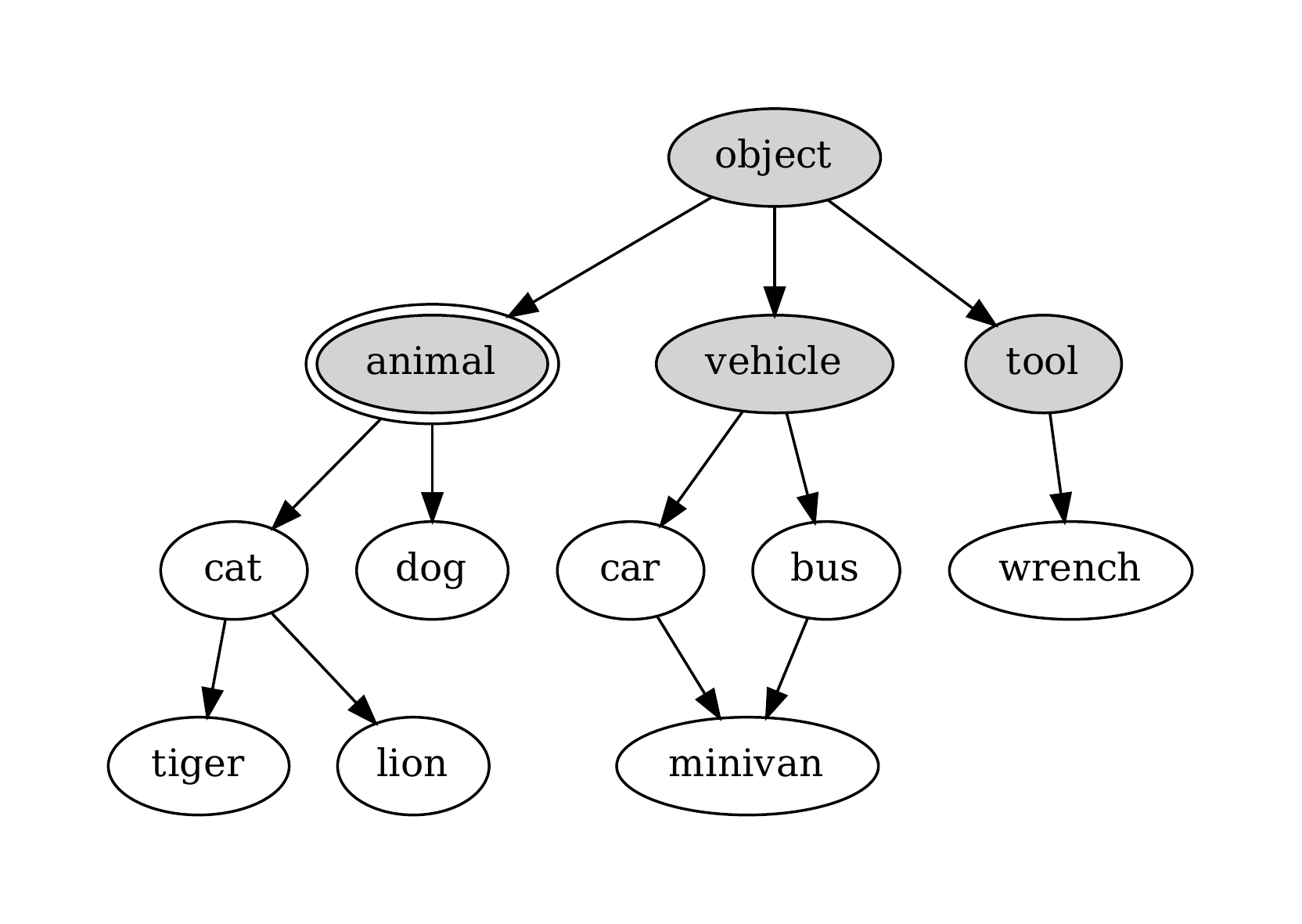}
		\label{fig:mdlmaskmod}
	}
	\caption{Comparison of original loss mask $m$ \cite{brust_integrating_2019} and
		our modification. In this example, the ground truth is \enquote{animal}. Dark nodes are trained and light nodes are not.}
	\label{fig:lossmasks}
\end{figure*}

In this work, we adapt a probabilistic hierarchical classifier proposed by Brust \& Denzler in \cite{brust_integrating_2019}.
It models the semantics associated with \textbf{is-a} relations using the following equation (cf.~\cite[eq.~(5)]{brust_integrating_2019}):
\begin{equation}
	P(Y_s^+|X) = P(Y_s^+|X, Y_{S'}^+)P(Y_{S'}^+|X),
\end{equation}
where $P(Y_s^+|X)$ is the probability of a label $s$ being present given an image $X$.
The event $Y_{S'}^+$ indicates that \emph{any} of the hypernyms of $s$ are present.
If no hypernyms of $s$ are present, then $s$ cannot be present itself,
except for the root.
This equation is recursive, with (cf.~\cite[eq.~(6)]{brust_integrating_2019}):
\begin{equation}
	P(Y_{S'}^+|X) = 1 - \prod_{i=1}^{|S'|}{1-P(Y_{s'_i}^+|X)},
\end{equation}
where $S'$ is the set of hypernyms of $s$.

To apply this model to a neural network, each label (precise and imprecise) is assigned one node in a sigmoid output layer.
A node assigned to $s$ predicts the conditional probability $P(Y_s^+|X, Y_{S'}^+)$.
The targets for each predictor are calculated by a label encoding $e$, which is 1 if a label is present and 0 otherwise.
Binary cross-entropy is used as a loss function.
To train nodes only when the truth value of their respective label is known, a loss mask $m$ is applied~\cite{brust_integrating_2019}.
Its precise definition, and our proposed modification to it, is given in the following.

\subsection{A Loss Mask for Semantically Imprecise Data}
The aforementioned classifier is in principle capable of handling semantically imprecise data.
However, it treats inner nodes as mutually exclusive with their children, assuming that a label is always as precise as possible.
A label \enquote{bird} then means \enquote{a type of bird, but not one present in the hierarchy}.
This is a result of optimizing for a slightly different task than ours.
For CHILLAX, we require \enquote{bird} to only mean \enquote{some type of bird}.

To achieve this, we modify the loss mask $m$ in eq. (12) of \cite{brust_integrating_2019} such that the children of the labeled node are not trained at all.
The original $m: \mathcal{Y}^+ \to \{0,1\}^{|\mathcal{Y}^+|} $ is defined in \cite{brust_integrating_2019} as:
\begin{equation}
	m(y)_s = \begin{cases}
		1 & \text{$y=s$ or}                                                    \\
		  & \exists (s,s') \in h: y = s' \text{ or }(y,s') \in \mathcal{T}(h), \\
		0 & \text{otherwise}, 
	\end{cases}
\end{equation}
where $s \in \mathcal{Y}^+$ is the component deciding whether the node representing $s$ should be trained, given a label $y$.
$h$ is the hyponymy relation and $\mathcal{T}(h)$ its transitive closure, such that $(s,s')\in h$ means $s$ \textbf{is-a} $s'$ \cite{brust_integrating_2019}.

In other words: given the label $y$, train the $s$ node if:
\begin{enumerate}
	\item $y = s$,
	\item $s$ \textbf{is-a} $y$, or
	\item $s$ \textbf{is-a} any (transitive) hypernym of $y$.
\end{enumerate}

To make the classifier compatible with our interpretation of imprecise labels, we need to remove the second criterion.
This way, the nodes representing the ground truth label's children are not trained, because we do not know their truth values.
We then define our alternative loss mask $m'$ as:
\begin{equation}
	m'(y)_s = \begin{cases}
		1 & \text{$y=s$ or}                                  \\
		  & \exists (s,s') \in h: (y,s') \in \mathcal{T}(h), \\
		0 & \text{otherwise},
	\end{cases}
\end{equation}
removing the second criterion $\exists (s,s') \in h: y = s'$ and only training the nodes where our knowledge is sufficient.
See \cref{fig:lossmasks} for a visualization of the original loss mask and our modification.

Note that changing the encoding $e$ is not necessary. Originally, it encodes children of the ground truth label as negative examples. This assignment does not fit our task, in which we do not know the truth value at all. However, our modified loss mask $m'$ ensures the relevant nodes are not trained either way. Only leaf nodes are evaluated for prediction.

\section{Modeling Semantic Imprecision}
\label{sec:noisemodels}

\begin{figure*}[t]
	\centering
	\subfloat [Web Crawling] {%
		\includegraphics[width=0.23\textwidth]{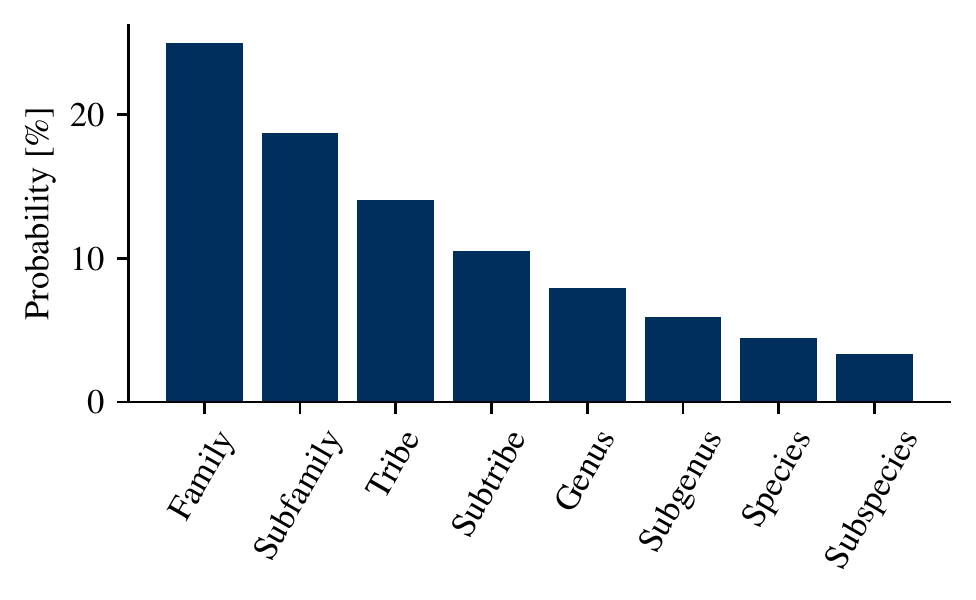}
		\label{sfig:geometric}
	}
	\subfloat[Volunteer] {%
		\includegraphics[width=0.23\textwidth]{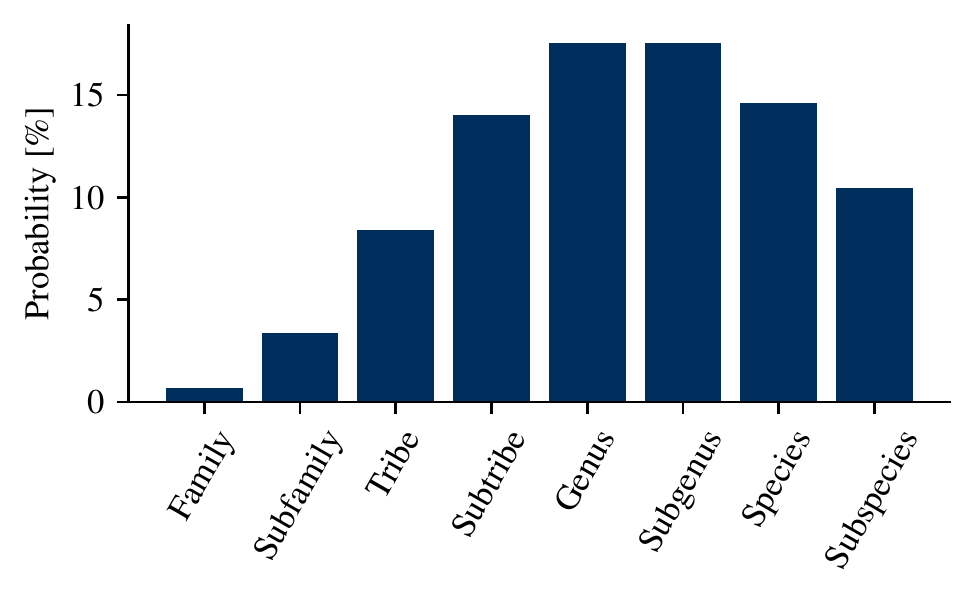}
		\label{sfig:poisson}
	}
	\subfloat[Relabeling \cite{deng_large-scale_2014}] {%
		\includegraphics[width=0.23\textwidth]{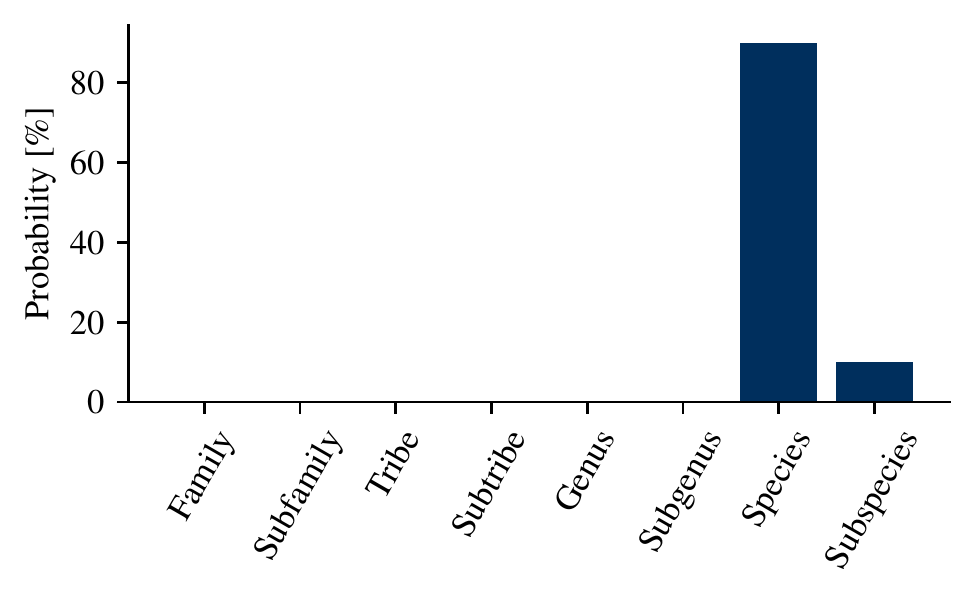}
		\label{sfig:deng}
	}
	\subfloat[Professional / Benchmark] {%
		\includegraphics[width=0.23\textwidth]{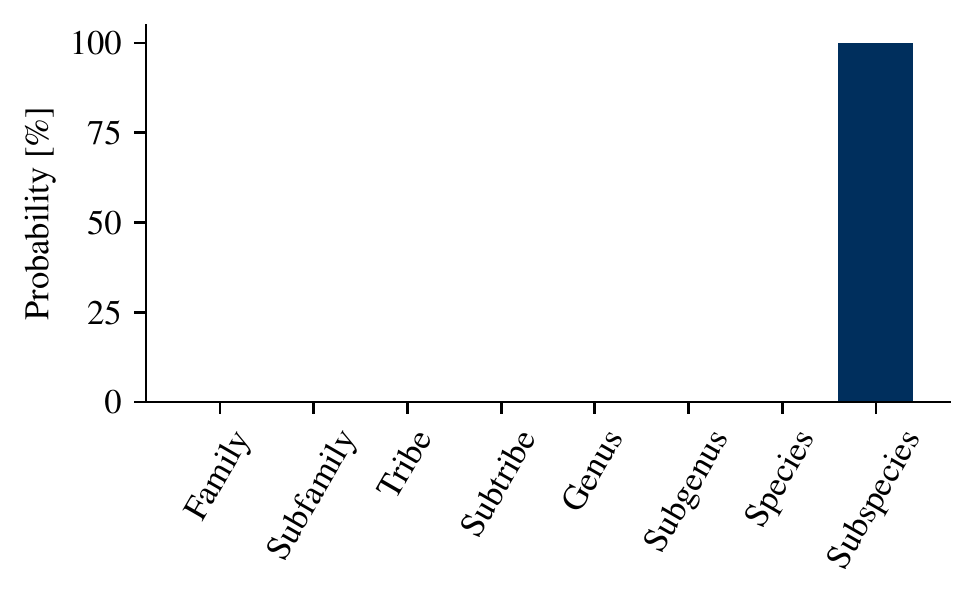}
		\label{sfig:bench}
	}
	\caption{The depth of a label in a given class hierarchy represents its precision. The graphs show distributions over label depth as modeled for different annotation sources. Benchmark data does not show any imprecision.}
	\label{fig:example-distributions}
\end{figure*}

We need a more fine-grained approach to label precision in order to study different types of semantically imprecise data properly.
Instead of classifying labels as either precise or imprecise, we define precision as depth in a given class hierarchy.

With this definition, we can model the precision statistics of different data sources.
In the following, we consider data crawled from the web as well as data provided by volunteers.
For reasons explained in the following, both are semantically imprecise.
We contrast these sources with benchmark data that shows no imprecision at all, and related work that models very mild imprecision \cite{deng_large-scale_2014}.

The models are visualized in \cref{fig:example-distributions} and validated on real-world data in \cref{ssec:realworlddata}.

\subsection{Web Crawling}
\label{ssec:webcrawling}
Labels crawled from the web are typically extracted from image captions, hashtags, or accompanying text.
The authors of this information are not labeling purposely, but instead attempting to describe an image for accessibility or encourage sharing and interaction.
Such a process seldom results in a precise label.
In fact, we expect most images to be barely labeled at all:
a caption of \enquote{this beautiful \#bird i saw while jogging} is more likely than \enquote{enjoy this non-breeding bunting!}
There are many more problems with web crawling, such as filtering, or a choice of pre-defined possible labels.
We address some of these challenges in \cref{ssec:realworlddata}, where we validate this model with real-life data.

We model the probability of a label having a certain depth in the hierarchy.
For web-crawled data, we apply a geometric distribution (\cref{sfig:geometric}).
This model is described, but not investigated further in \cite{deng_large-scale_2014}.

Depending on the class hierarchy, it can be useful to offset the distribution by a number of layers.
The offset is such that the first layers have zero probability of occurring, and the distribution is \enquote{shifted to the right}.
This shift models the extreme degree of abstractness in the first layers of hierarchies such as WordNet \cite{fellbaum_wordnet_1998},
which is not useful in practice.
In \cref{fig:realworld}, this shift can be observed.

\subsection{Volunteer Labelers}
\label{ssec:volunteer}
We consider images labeled by volunteers, or for small amounts of money, \eg Mechanical Turkers.
Volunteers are aware that they are labeling and annotate to the best of their abilities, but may still not be experts in the problem field.
Only a small fraction can label at the most precise level, but most will be able to provide some information.
Compared to web crawling, it is easier to restrict volunteer annotators to a specific set of labels and images.
This restriction eliminates the problem of mapping annotations to a class hierarchy.
There is also no need to filter the data if images are pre-selected.

How to perform interaction in a manner that results in the highest possible precision is still an open research question.

The combination of a best-effort annotation on the one hand and lack of expertise on the other leads us to assume a Poisson-distributed dataset (\cref{sfig:poisson}).
Shifting the distribution depending on the class hierarchy in question can be useful, as described in \cref{ssec:webcrawling}.

\subsection{Relabeling to Immediate Parents}
\label{ssec:dengmodel}
Deng \etal initially propose this type of noise in \cite{deng_large-scale_2014}.
They use it to evaluate their method HEX (Hierarchy and Exclusion graphs).
Their setup replaces a fixed fraction of precise labels with the direct parent label (cf. \cref{sfig:deng}).
In a deep hierarchy, the precision is only marginally affected.
There is not much information lost by going up only one level,
particularly in a deep hierarchy with low fan-out.
Determining a single parent as replacement is only possible in tree hierarchies, but not all directed acyclic graphs.

\subsection{Professional / Benchmark Datasets}
Professionally created datasets exhibit few or none of the quality problems described in the previous sections.
They are typically acquired by asking paid domain experts for labels, or by running extensive citizen science projects.
Before labeling, the images are carefully selected and controlled for quality.
Multiple annotators label the same image to further improve accuracy by consensus.
This process is the gold standard, where labels are only of the highest precision and accuracy.
However, only a small number of these datasets exist and are available to the general public.
For budget reasons, a high-quality dataset may not be the first choice for many projects.

\Cref{sfig:bench} illustrates this extreme situation.

\section{Experiments and Evaluation}
In this section, we validate our proposed method CHILLAX in three experiments.
We first investigate the effects of imprecision using the models proposed in \cref{sec:noisemodels} in detail.
This controlled experiment is important to isolate the effects.
Then, we add inaccuracy, \ie incorrect labels, to better reflect a real-world setting, where both types of label noise will occur.
We also compare CHILLAX with HEX \cite{deng_large-scale_2014} to validate its competitiveness.

\subsection{Setup}
\label{ssec:expsetup}
This section details our experimental setup, including datasets, methods and technical details.
We use the CHIA framework for our experiments. Code
\footnote{\url{https://github.com/cvjena/chia} and \url{https://github.com/cvjena/chillax}}
is publicly available.

\paragraph{Training Data}
There are no publicly available benchmarks specifically for this use case.
Instead, we build synthetic datasets and carefully validate the underlying assumptions.
The North American Birds (NABirds,~\cite{van_horn_building_2015}) and ImageNet Large Scale Visual Recognition Challenge 2012 (ILSVRC2012,~\cite{russakovsky_imagenet_2015}) datasets are used as a starting point.
We degrade the training datasets using the noise models described in \cref{sec:noisemodels}.
The hierarchies that are supplied as part of the datasets are used in all cases to avoid any manual mapping errors.
NABirds offers a tree hierarchy. WordNet, underlying ILSVRC2012, is not a tree as it contains (undirected) cycles.

Our goal is to modify the training datasets such that the distribution over label depths in the hierarchy matches a given noise model.
For each original label, we sample a depth from the model and go up the hierarchy until this depth is reached.
If there are multiple parents to choose from at the targeted depth, one is randomly selected.
There is never a need to go down the hierarchy,
since all datasets are labeled as leaf nodes.
To fit each dataset, all distributions are truncated to not exceed the maximum depth of the given hierarchy.
We apply no shift to the distributions in our synthetic data.

There is a distinct advantage to using synthetic data.
It allows us to control the exact degree of imprecision, highlighting the strengths and weaknesses of each method.

\paragraph{Baselines}
We evaluate our method against strong baselines, using the typical softmax classifier and cross-entropy loss function.
The softmax classifier assumes mutually exclusive classes.
As such, this classifier can only learn from examples labeled as leaf nodes.
However, the degradation process can remove a large fraction of these precise examples.
The resulting imprecise examples cannot be used directly by the softmax classifier.

We propose two baseline strategies for handling the imprecise examples:
\begin{enumerate}
	\item Leaves only: filter out imprecise examples, leaving only a portion of the dataset.
	\item Random leaf: Project imprecise examples to a random descendant leaf node, utilizing the whole dataset, but introducing inaccuracies.
\end{enumerate}

These baseline methods represent the current state of not using imprecise data at all, and the alternative of making an educated guess and turning a weak label intro a strong (precise) one.
Without imprecision, they are identical.
To the best of our knowledge, the only other method capable of performing our task is HEX \cite{deng_large-scale_2014}. However, it cannot be replicated or applied to new datasets with reasonable effort.
We evaluate our method against HEX in \cref{ssec:hex}.
A typical hierarchical classifier would need considerable modifications because of our task's asymmetry of training and prediction label sets.
Since these modifications (specifically the constraint on predicted labels) are not trivial, but have a large influence on the results, we do not use common hierarchical classifiers as baselines.

For better comparison with other works, we report metrics on the respective unmodified validation datasets.

\paragraph{Neural Network Training}
Our deep learning setup matches the ResNet50 described in \cite{brust_integrating_2019}, with a few exceptions.
Our learning rate schedule is SGDR \cite{loshchilov_sgdr_2017}.
There is no data augmentation except for random horizontal flipping.
For ILSVRC2012, our main goal to approximate the performance of \cite{deng_large-scale_2014} when there is no noise,
while our NABirds setup should come closer to the state of the art.
A detailed listing of hyperparameters can be found in \cref{tbl:hyperparams}.

\begin{table}
	\caption{Detailed Hyperparameters for all Experiments.}
	\label{tbl:hyperparams}
	\centering
	\footnotesize{%
		\begin{tabular}{lll}
			\toprule
			Setting            & NABirds                                     & ILSVRC2012         \\
			\midrule
			Max. Learning Rate & 0.003                                       & 0.2                \\
			Min. Learning Rate & $10^{-6}$                                   & $10^{-5}$          \\
			$T_0$              & 80 epochs                                   & 10 epochs          \\
			Warmup             & 1 epoch @ lr=0.01                           & 5 epochs @ lr=0.05 \\
			Training Duration  & 80 epochs                                   & 20 epochs          \\
			Pre-Training       & iNaturalist 2018 \cite{Cui2018iNatTransfer} & ---                \\
			Batch Size         & 32                                          & 128                \\
			Image Size         & $512\times512$                              & $256\times256$     \\
			Random Crop Size   & $448\times448$                              & $224\times224$     \\
			\bottomrule
		\end{tabular}}
\end{table}

\subsection{Imprecision Only}
\label{ssec:expimponly}
We first investigate the effects of imprecise labels in a controlled fashion.
Only the noise model is applied, and there are no other modifications to the training data.
\Cref{tbl:nabirds_accuracy_imprecision_only} shows the results obtained by six runs each of our method CHILLAX and the two baselines.

We observe that CHILLAX's performance relative to the baselines increases with the level of noise applied to the training data.
Without noise, it reaches an accuracy of $81.4\% \pm 0.2$, thus performing slightly worse than the baseline one-hot classifier at $82.8\% \pm 0.2$.

But adding noise, we outperform the baselines in almost all cases, except for the geometric noise model with $q > 0.9$.
We observe substantial improvements in high noise situations such as the Poisson model at $\lambda=1$.

For geometric and Poisson noise, the \enquote{leaves only} baseline shows better results than the \enquote{random leaf} variant.
This is likely because of the strong confusions caused by selecting a random leaf.
The results in the relabeling scenario support this.
Here, selecting a random leaf is beneficial because it is not as likely to be false as in the other scenarios.
From the second to last level in the hierarchy, there are only approximately two child nodes to choose from, instead of the tens or hundreds that are reachable from higher levels, as happens in the geometric and Poisson cases.

The slightly worse accuracy without any noise is possibly due to problematic assumptions about the correlation between visual and semantic similarity \cite{brust_not_2018,brust_integrating_2019}.
This affects CHILLAX negatively, because it is based on hierarchical classification.
But this effect is outweighed by the advantages of learning directly from imprecise data in almost all cases.

\begin{table}
	\centering
	\caption{Accuracy (\%) and standard deviation on the NABirds dataset of our method CHILLAX and the baselines. We evaluate the geometric (web crawling), Poisson (volunteer), and relabeling imprecision noise models. There is no added inaccuracy.}
	\footnotesize{
\begin{tabular}{lllll}
    \toprule
    Geometric &   $50\%$ &   $80\%$ &  $90\%$ & $95\%$ \\
    \midrule
    BL: leaves only &  42.9 $\pm$ 0.6 &  75.2 $\pm$ 0.3 &  \textbf{79.6} $\pm$ 0.1 &  \textbf{81.4} $\pm$ 0.2 \\
    BL: random leaf &  12.5 $\pm$ 0.4 &  51.4 $\pm$ 0.5 &  67.9 $\pm$ 0.2 &  75.5 $\pm$ 0.3 \\
    CHILLAX (Ours)                   &  \textbf{48.9} $\pm$ 1.0 &  \textbf{75.6} $\pm$ 0.1 &  79.1 $\pm$ 0.3 &  80.3 $\pm$ 0.1 \\
    \midrule
    Poisson &   $\lambda=1$ &     $\lambda=2$ &     $\lambda=3$ &     $\lambda=4$ \\
    \midrule
    BL: leaves only &  26.5 $\pm$ 0.8 &  61.9 $\pm$ 0.5 &  74.9 $\pm$ 0.3 &  79.1 $\pm$ 0.2 \\
    BL: random leaf &  11.1 $\pm$ 0.4 &  36.8 $\pm$ 0.4 &  59.0 $\pm$ 0.5 &  70.6 $\pm$ 0.3 \\
    CHILLAX (Ours)                  &  \textbf{42.9} $\pm$ 0.4 &  \textbf{70.1} $\pm$ 0.2 &  \textbf{77.7} $\pm$ 0.3 & \textbf{ 80.1} $\pm$ 0.1 \\
    \midrule
    Relabeling \cite{deng_large-scale_2014} &   $99\%$ & $95\%$ &  $90\%$ &  $50\%$ \\
    \midrule
    BL: leaves only &   9.0 $\pm$ 0.6 &  28.2 $\pm$ 0.6 &  45.6 $\pm$ 1.6 &  76.3 $\pm$ 0.3 \\
    BL: random leaf &  60.9 $\pm$ 0.6 &  61.5 $\pm$ 0.5 &  63.5 $\pm$ 0.2 &  74.3 $\pm$ 0.5 \\
    CHILLAX (Ours)                  &  \textbf{63.2} $\pm$ 0.4 &  \textbf{70.4} $\pm$ 0.7 &  \textbf{75.1} $\pm$ 0.4 &  \textbf{80.9} $\pm$ 0.1 \\
    \bottomrule
\end{tabular}
}
	\label{tbl:nabirds_accuracy_imprecision_only}
\end{table}

\subsection{Semantic Error Analysis}
\label{ssec:lcaanalysis}
Hierarchical classifiers often exhibit a different quality of errors from regular ones, \ie they make \enquote{better} mistakes \cite{bertinetto_making_2019}.
To determine if CHILLAX also benefits from this property, we calculate the depth of the lowest common ancestor (LCA) of each mispredicted label-prediction pair in the validation set.

\Cref{tbl:lcad} shows the LCA depth corresponding to the previously reported results, obtained during the same experiment (see \cref{ssec:expimponly}).
In most cases, the baselines behave as expected.
The LCA depth increases when the labels are more precise, except for the random leaf baseline in the relabeling setting.
There, the LCA depth decreases with more precise labels.
In other words, the baseline makes \enquote{better} mistakes when trained on noisy data.
CHILLAX shows the same unexpected behavior, where the least precise labels always lead to the highest LCA depth.

A possible explanation is that the classifier \enquote{focuses} on the earlier layers of the class hierarchy, simply because the respective labels make up most of the training data in the noisier case.
However, this does not apply to the relabeling case, where label depth is reduced by one layer at most, and we still observe similar results.

Regardless of this anomaly, in most cases, CHILLAX makes the \enquote{best} mistakes from a semantic perspective.


\begin{table}
	\centering
	\footnotesize{%
		\caption{Lowest common ancestor (LCA) depth between ground truth and mispredictions on the NABirds dataset. We evaluate the geometric (web crawling), Poisson (volunteer), and relabeling imprecision noise models. \emph{Higher is better}.}
		\label{tbl:lcad}
		\begin{tabular}{lllll}
    \toprule
    Geometric &   $50\%$ &   $80\%$ &  $90\%$ & $95\%$ \\
    \midrule
    BL: leaves only &          2.00 &          2.14 &          2.17 &           2.18 \\
    BL: random leaf &          1.78 &          2.07 &          2.13 &           2.16 \\
    CHILLAX (Ours)                  &          \textbf{2.39} &          \textbf{2.27} &          \textbf{2.26} &           \textbf{2.24} \\
    \midrule
    Poisson &   $\lambda=1$ &     $\lambda=2$ &     $\lambda=3$ &     $\lambda=4$ \\
    \midrule
    BL: leaves only &        1.85 &        2.10 &        2.14 &        2.17 \\
    BL: random leaf &        1.89 &        2.21 &        2.29 &        2.27 \\
    CHILLAX (Ours)                  &       \textbf{2.51} &        \textbf{2.39} &        \textbf{2.31} &        2.27 \\
    \midrule
    Relabeling \cite{deng_large-scale_2014} &   $99\%$ & $95\%$ &  $90\%$ &  $50\%$ \\
    \midrule
    BL: leaves only &          1.74 &          1.90 &         1.97 &         2.13 \\
    BL: random leaf &          2.79 &          \textbf{2.76} &         \textbf{2.73} &         \textbf{2.46} \\
    CHILLAX (Ours)                  &          \textbf{2.83} &          2.68 &         2.52 &         2.28 \\
    \bottomrule
\end{tabular}
}
\end{table}

\subsection{Inaccuracy and Imprecision}
While the previous experiment is important to evaluate our method in a controlled manner, we should also investigate the effects of inaccurate data.
When acquiring labeled data from noisy sources, it is unlikely that it is only imprecise.
Literature predominantly considers inaccuracy when mentioning label noise (see \cref{sec:relatedwork}).
Inaccuracy can result from a lack of knowledge, errors language processing used in web crawling, and can also be intentional as a form of vandalism.

To simulate inaccuracy in this experiment, we replace a fraction of the precise training data with randomly selected labels of the same high precision, \ie leaf nodes.
Imprecision is applied afterward, as described in \cref{ssec:expsetup}, thus compounding both types of label noise.
Note that the final fraction of inaccurate annotations can be lower because two confused classes can have overlapping ancestors that are selected by the imprecision process.
Alternatively, one could apply imprecision first. However, applying inaccuracy afterwards is not trivial since there are more options for confusion that can not be treated equally.

\Cref{tbl:nabirds_accuracy_twonoise} shows the effects of randomly confusing $10\%$ of the training data in six runs each.
Without any additional imprecision, the baseline reaches an accuracy of $77.3\% \pm 0.1$, slightly outperforming our method at $75.3\% \pm 0.4$.
Applying the imprecision models, the results are similar to \cref{ssec:expimponly}.
We still observe CHILLAX performing better than the baselines when there is more imprecision, although the amount required for outperformance is slightly elevated.
A $1\%$ confused training data experiment exhibits similar behavior.

CHILLAX is affected disproportionately by inaccuracy label noise.
When the baseline classifier encounters two randomly confused classes, two nodes are trained in the wrong direction.
Applying the same confusion to a hierarchical classifier, two whole paths can be affected.
A wrong decision in an early layer causes mispredictions on a much larger scale.

Despite these fundamental disadvantages of hierarchical classifiers in the face of a high fraction of inaccurate labels, CHILLAX performs very well, and still better than both baselines in most cases.

\begin{table}
	\centering
	\caption{Accuracy (\%) and standard deviation on the NABirds dataset of our method CHILLAX and the baselines. We evaluate the geometric (web crawling), Poisson (volunteer), and relabeling imprecision noise models. $10\%$ inaccuracy is applied.}
	\footnotesize{
\begin{tabular}{lllll}
    \toprule
    Geometric &   $50\%$ &   $80\%$ &  $90\%$ & $95\%$ \\
    \midrule
	BL: leaves only &  36.4 $\pm$ 1.0 &  \textbf{69.2} $\pm$ 0.1 &  \textbf{73.7} $\pm$ 0.2 &  \textbf{75.8} $\pm$ 0.3 \\
	BL: random leaf &   9.4 $\pm$ 0.2 &  46.6 $\pm$ 0.6 &  62.6 $\pm$ 0.1 &  69.9 $\pm$ 0.2 \\
	CHILLAX (Ours)                  &  \textbf{38.7} $\pm$ 0.7 &  67.3 $\pm$ 0.2 &  72.1 $\pm$ 0.1 &  73.5 $\pm$ 0.2 \\
    \midrule
    Poisson &   $\lambda=1$ &     $\lambda=2$ &     $\lambda=3$ &     $\lambda=4$ \\
    \midrule
	BL: leaves only &  22.1 $\pm$ 0.4 &  54.4 $\pm$ 1.2 &  67.9 $\pm$ 0.1 &  \textbf{73.1} $\pm$ 0.6 \\
	BL: random leaf &  10.0 $\pm$ 0.3 &  33.1 $\pm$ 0.6 &  53.1 $\pm$ 0.7 &  65.4 $\pm$ 0.2 \\
	CHILLAX (Ours)                  &  \textbf{34.6} $\pm$ 1.2 &  \textbf{60.5} $\pm$ 0.3 &  \textbf{69.8} $\pm$ 0.3 &  72.8 $\pm$ 0.2 \\
    \midrule
    Relabeling \cite{deng_large-scale_2014} &   $99\%$ & $95\%$ &  $90\%$ &  $50\%$ \\
    \midrule
	BL: leaves only &   8.2 $\pm$ 0.6 &  23.5 $\pm$ 1.1 &  39.1 $\pm$ 1.6 &  70.5 $\pm$ 0.7 \\
	BL: random leaf &  57.2 $\pm$ 0.0 &  57.7 $\pm$ 0.1 &  59.2 $\pm$ 0.1 &  69.2 $\pm$ 0.4 \\
	CHILLAX (Ours)                  &  \textbf{58.3} $\pm$ 0.7 &  \textbf{63.6} $\pm$ 0.5 &  \textbf{67.8} $\pm$ 0.4 &  \textbf{74.3} $\pm$ 0.5 \\
    \bottomrule
\end{tabular}
}
	\label{tbl:nabirds_accuracy_twonoise}
\end{table}

\subsection{Comparison to HEX \cite{deng_large-scale_2014}}
\label{ssec:hex}

\begin{figure*}
	\centering
	\includegraphics[width=\textwidth]{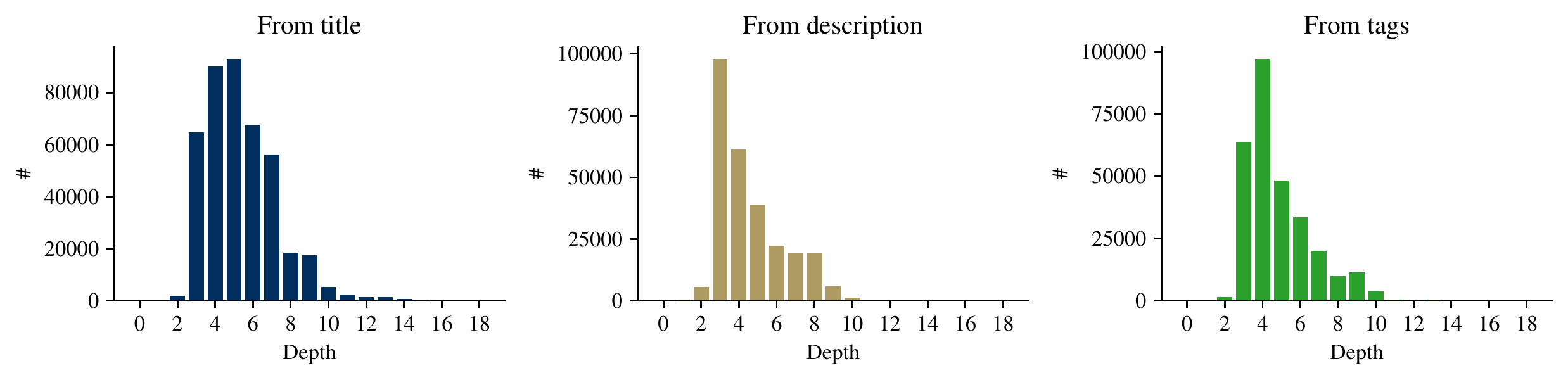}
	\caption{Imprecision in Flickr: Frequency over depth in the WordNet hierarchy. A most descriptive synset is calculated per image, and separately for the title, description, and tags in the image metadata to highlight the differences.}
	\label{fig:realworld}
\end{figure*}

We do not consider the experimental protocol proposed by Deng \etal in \cite{deng_large-scale_2014} to be representative of our specific task.
Relabeling classes to their immediate parents in a deep hierarchy does not reduce label precision substantially.
This effect can be observed in \cref{ssec:expimponly}, where selecting a random leaf node achieves good performance even when relabeling almost all samples to their parents.
We also visualize the small influence on precision in \cref{fig:example-distributions} (c) and (d).

Still, a comparison with existing methods is an important part of validating a proposed method and the setup described in \cite{deng_large-scale_2014} comes closest to our scenario.
As such, their method HEX can be considered the state of the art.

For a fair comparison, we first tune CHILLAX on ILSVRC2012 to achieve similar performance to HEX when there is no noise.
CHILLAX reaches an accuracy of $62.5\%$ ($83.5\%$ top-5) on the validation set, compared to $62.6\%$ ($84.3\%$ top-5).
This setup means that our method does not have any advantage simply because of a more recent deep neural network architecture, so any differences in performance can be attributed to the different hierarchical classifiers more easily.
We observe the performance of CHILLAX and both baselines under the four noise levels proposed by Deng \etal and compare these results to their HEX.

\begin{table}
	\centering
	\footnotesize{%
		\caption{Top-1 (top-5) accuracy ($\%$) on ILSVRC2012 with relabeling to immediate parent. We evaluate our method CHILLAX against HEX \cite{deng_large-scale_2014} following the same experimental protocol.}
\begin{tabular}{lllll}
    \toprule
    Method $\backslash$ Frac. &   $99\%$ & $95\%$ &  $90\%$ &  $50\%$ \\
    \midrule
    BL: leaves only &    6.9 (16.4) &    30.2 (51.6)&  41.6 (64.0) &         60.2 (82.2) \\
    BL: random leaf &   32.1 (68.1) &   35.7 (71.8) &  37.9 (74.2) &   56.0 (81.9) \\
    \midrule
	HEX \cite{deng_large-scale_2014} & \textbf{41.5} (68.5) & \textbf{52.4} (77.2) & 55.3 (79.4) & 58.2 (80.8)\\
	CHILLAX (Ours)        & 38.1 (\textbf{68.6}) &   52.1 (\textbf{78.1}) &\textbf{ 55.5 (80.2)} & \textbf{62.1 (83.6)}\\
    \bottomrule
\end{tabular}
}
	\label{tbl:deng}
\end{table}

\Cref{tbl:deng} shows the results of this experiment. At $50\%$ to $90\%$ noise, CHILLAX outperforms HEX in both metrics.
When inspecting top-5 accuracy, it is superior at all noise levels.
However, HEX does achieve a higher top-1 accuracy at $95\%$ and $99\%$.
Both baselines are not competitive in any setting, although selecting random leaves performs surprisingly well at high relabeling fractions.

The comparison between CHILLAX and HEX does not correspond to our earlier findings where our method's performance advantage over the baselines increased with higher noise.
However, HEX is a hierarchical method like ours and very different from the baselines in the previous experiment.

The advantage in top-5 accuracy, even at lower top-1 accuracy, comes from our method of computing per-class probabilities.
The conditional representation means that each predictor can concentrate independently on a straightforward task and does not need to coordinate with other predictors like in a softmax setting.
Sometimes, this missing global view of classification results in a wrong ranking of predictions because predictors are not trained to \enquote{overpower} each other.
This results in a slightly lower top-5 accuracy.
But it increases the probability of predicting less represented classes, leading to higher top-5 accuracy.

Overall, CHILLAX is competitive with HEX, but more generally applicable to data with arbitrary levels of imprecision.

\subsection{Imprecision in Real-World Datasets}
\label{ssec:realworlddata}

To validate the models described in \cref{sec:noisemodels}, we analyze image metadata from Flickr \footnote{\url{https://www.flickr.com/}} as an example.
We obtain approximately $1,500,000$ data points corresponding to images taken from January 1st, 2019 to December 31st, 2019.
Our goal is to map each image to oe synset in WordNet \cite{fellbaum_wordnet_1998} if at all possible.
Note that the first few layers of the WordNet noun hierarchy do not contain concepts that are meaningful descriptions of image content.
Hence, they rarely occur in the image metadata.
Thus, we expect the precision distributions to be slightly \enquote{shifted}.
We also expect a slight bias towards terms used in photography, such as \enquote{aperture}, \enquote{lens} and \enquote{focal length}.
The respective synsets have depths 7 to 10.

The title, description and list of (human-supplied) tags are processed separately for a more detailed analysis.
This information is tokenized and then lemmatized.
The lemmatization tries to find a matching noun because that is the most likely setup for a classification dataset.
For each lemma, there are many possible synsets.
We select the one closest to the root, avoiding over-interpretation.
The result of this process in a set of synsets per image.
To obtain a single label, the synset farthest from the root is selected because we expect it to be most descriptive of the image.
If a synset cannot be determined, we ignore the image altogether.

\Cref{fig:realworld} shows the distribution over depth in WordNet of the most descriptive synset per image.

The distribution obtained from the image title aligns most closely with the volunteer model as described in \cref{ssec:volunteer}.
For the other distributions, the web crawling model (\cref{ssec:webcrawling}) fits well when shifting the distribution to ignore the first three or four (meaningless) layers of the hierarchy.

Title, description and tags exhibit different characteristics because they are used towards contrasting goals.
Titles are a more artistic expression and very descriptive, while tags should be generic to match as many search queries as possible.
The description field rarely contains additional information, if used at all.
It is often a sentence combining the title and tags, or a direct copy of the title.
The distribution (\cref{fig:realworld}) appears to reflect this property.

We consider this real-world data a validation of the models proposed in \cref{sec:noisemodels}.

\section{Conclusion}
In this work, we consider the rarely discussed, but critical task of learning from semantically imprecise data.
Annotation processes such as web crawling or crowd sourcing naturally produce weak labels with imprecision, where we model precision in terms of depth in a given class hierarchy.

Conventional classifiers, \eg deep neural networks with softmax outputs learning from one-hot encoded labels, are incompatible with any but the most precise training data.
Because they model classes as mutually exclusive, they will draw wrong conclusions if they encounter imprecise labels.

Our method CHILLAX is capable of weak supervision, \ie learning from imprecise and precise labels at the same time.
It always extrapolates to predictions at the highest precision.

To the best of our knowledge, this learning problem is not considered in existing literature.
Only \cite{deng_large-scale_2014} describes a very basic formulation.
Consequently, there are no benchmarks for this problem.
To evaluate our method's performance, we build a synthetic dataset based on a popular benchmark.
We take the North American Birds dataset \cite{van_horn_building_2015} because it ships with a well understood class hierarchy.
Species recognition also represents a typical use of volunteer annotations, where imprecise labels will occur.
The training set is weakened by replacing labels with ancestors of a certain depth, according to our imprecision models (see \cref{sec:noisemodels}).

If synthetic data is to be representative of real-world problems, the underlying assumptions have to be carefully validated.
We do this by analyzing image metadata crawled from Flickr.
The distributions obtained from image titles, descriptions, and tags validate our model assumptions.

Using this real-world validated synthetic data, we show that our method outperforms strong baselines in almost all cases.
On imprecise NABirds modelling a noisy volunteer annotation scenario, CHILLAX achieves an accuracy of $42.9\%$, a substantial improvement to the best baseline at $26.5\%$.
We also compare our method to the state-of-the-art HEX \cite{deng_large-scale_2014} using their ILSVRC2012 experimental protocol and observe superior performance in most settings. At $50\%$ relabeling, CHILLAX reaches an accuracy of $62.1\%$ vs. HEX at $58.2\%$.

\textbf{Future Work} With higher imprecision, a significant amount of images are labeled as the root of a given class hierarchy.
If the fraction of such examples is high, an unsupervised learning method could help increase sample efficiency.

Building a new dataset from real sources is an interesting next step.
Examples of such data sources are discussed in  \cref{sec:noisemodels}.
Several small problems, including the development of an interaction mechanic for imprecise labeling, have to be solved along the way.

\textbf{Acknowledgments}
The computational experiments were performed on resources of Friedrich Schiller University Jena supported in part by DFG grants INST 275/334-1 FUGG and INST 275/363-1 FUGG. 


%

\bibliographystyle{IEEEtran}
\bibliography{paper}

\end{document}